\documentclass[10pt,twocolumn,letterpaper]{article}

\usepackage[pagenumbers]{cvpr}

\usepackage{cuted}

\usepackage{currfile}

\usepackage{caption}

\usepackage{xspace}
\newcommand{\method}{GuideCache\xspace}
\usepackage{amsmath,amssymb,multirow,colortbl}
\usepackage{iftex}
\ifXeTeX
  \usepackage[no-math]{fontspec}
  \IfFontExistsTF{SimSun}{\newfontfamily\zhfont{SimSun}}{\newfontfamily\zhfont{simsun.ttc}[Path=C:/Windows/Fonts/]}
  
  \normalfont
\else
  \usepackage{CJKutf8}
  
\fi

\definecolor{cvprblue}{rgb}{0.21,0.49,0.74}
\usepackage{stfloats}
\usepackage[pagebackref,breaklinks,colorlinks,allcolors=cvprblue]{hyperref}

\def\confName{CVPR}
\def\confYear{2027}

\title{Text Lives in the Guidance: Preventing Silent Text Rewriting in Cached Diffusion Transformers}

\author{Kun~Zhou\textsuperscript{1,2}\quad Wenmian~Yang\textsuperscript{3}\quad Zhensheng~Wang\textsuperscript{1,2}\quad Weijia~Jia\textsuperscript{2}\\[0.35em]
{\small \textsuperscript{1}School of Artificial Intelligence, Beijing Normal University, Beijing 100875, China}\\
{\small \textsuperscript{2}Institute of Artificial Intelligence and Future Networks, Beijing Normal University, Zhuhai 519087, China}\\
{\small \textsuperscript{3}School of Computer Science and Technology, Beijing Jiaotong University, Beijing 100044, China}}
\hypersetup{pdftitle={Text Lives in the Guidance: Preventing Silent Text Rewriting in Cached Diffusion Transformers}, pdfauthor={Kun Zhou, Wenmian Yang, Zhensheng Wang, Weijia Jia}}

\begin{document}
\twocolumn[{\renewcommand\twocolumn[1][]{#1}\maketitle
\begin{center}
\centering
\captionsetup{type=figure}
\includegraphics[width=\textwidth]{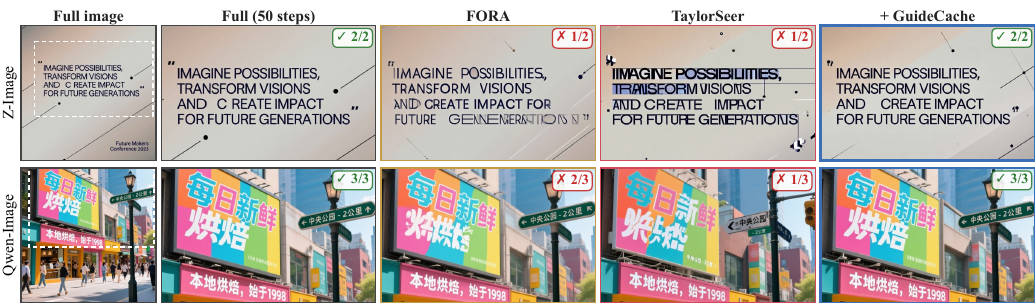}
\captionof{figure}{\textbf{Feature caching silently rewrites text.} Full-compute images (left) and text crops on Z-Image (top) and Qwen-Image (bottom); tags count the target strings that OCR reads correctly. FORA and TaylorSeer (2.94$\times$ fewer FLOPs) keep the layout of full compute but garble the text. With \method (2.41$\times$ and 2.46$\times$), TaylorSeer writes it correctly.}
\label{fig:teaser}
\end{center}
\vspace{0.4em}
}]
\begin{abstract}
Diffusion transformers (DiTs) now render long, legible text in images, and feature caching is a standard way to accelerate them.
Caches are judged by image-level metrics, which do not check whether the rendered text is still correct.
We find that caching silently rewrites this text.
On Z-Image, TaylorSeer at 2.9$\times$ fewer FLOPs lowers word accuracy from 62\% to 46\%, while ImageReward and CLIP Score barely change.
Sampling with fewer steps at the same compute keeps almost all of this text, so the loss comes from how the cache predicts the skipped steps.
Decomposing the cache error along classifier-free guidance (CFG), we trace this loss mainly to the guidance term, the difference between the conditional and unconditional predictions, which caches extrapolate worst on text.
Correcting the guidance inside the text alone recovers most of the lost text.
We therefore propose \method, which recomputes both CFG branches only on the tokens with the largest predicted guidance error, using the guidance terms that the cache already stores.
At 2.4$\times$ fewer FLOPs, \method restores full-compute prompt fidelity on Z-Image and Qwen-Image, and on Z-Image it renders text as well as or better than ToCa, TeaCache, and FasterCache while staying much closer to the full model's output.
\method works with caches of any order, and on top of fewer sampling steps it keeps more text than step reduction at a higher speedup.
These results suggest that acceleration methods for text-rendering models should be evaluated on the text they render.
\end{abstract}

\section{Introduction}
\label{sec:intro}

Text-to-image models have long struggled to render legible text.
Recent diffusion transformers (DiTs), including Qwen-Image~\cite{wu2025qwenimage} and Z-Image~\cite{zimage2025}, now generate long multilingual text on posters and product packaging.
This capability supports applications in design and e-commerce, where a single misspelled word can make an otherwise convincing image unusable.
It also places a strict requirement on acceleration.
A small change to a stroke can alter the identity of a character even when the surrounding image looks unchanged.

Generating these images remains expensive.
A $1024^2$ image requires dozens of passes through a multi-billion-parameter network, and classifier-free guidance (CFG)~\cite{ho2022cfg} doubles the cost of each pass.
Feature caching exploits the slow change of features across timesteps to reduce this cost.
It evaluates the full network on selected anchor steps and reuses~\cite{selvaraju2024fora}, extrapolates~\cite{liu2025taylorseer,cai2026lesa}, or partially recomputes~\cite{zou2024toca} cached features on the remaining steps.
These methods report little change in image-level metrics such as FID and ImageReward.
However, these metrics do not read the generated text, whose quality is illustrated only through qualitative examples in the caching literature.
Text-rendering benchmarks such as LongText-Bench~\cite{xomni2025} and CVTG-2K~\cite{cvtg2025} evaluate text with OCR, but at full compute.
Whether caching preserves this capability therefore requires direct evaluation.

We find that feature caching can preserve the image while changing its words.
As shown in \cref{fig:teaser}, accelerated outputs retain the layout and objects of full-compute images, yet render slogans and billboard text with misspelled or garbled characters.
We call this \emph{silent text rewriting}, because image-level scores respond far less than text scores.
To measure it, we pair accelerated and full-compute images generated with the same prompt and seed, and read their text with OCR.
Word accuracy and prompt fidelity measure agreement with the target strings, while box fidelity measures agreement with the full-compute text.
On LongText-Bench, TaylorSeer at 2.9$\times$ fewer FLOPs lowers Z-Image's word accuracy from 62\% to 46\%, while ImageReward and CLIP Score barely change.
Reducing the sampling steps at the same compute preserves almost all of this text.
On Qwen-Image, step reduction loses as much text as caching but visibly degrades the image, whereas the cache retains its overall appearance.
The loss occurs across cache orders, grows with extrapolation order and speedup, and is most severe on small text.
We therefore ask why a cache that preserves the image gets the text wrong.

\begin{figure}[t]
\centering
\includegraphics[width=\linewidth]{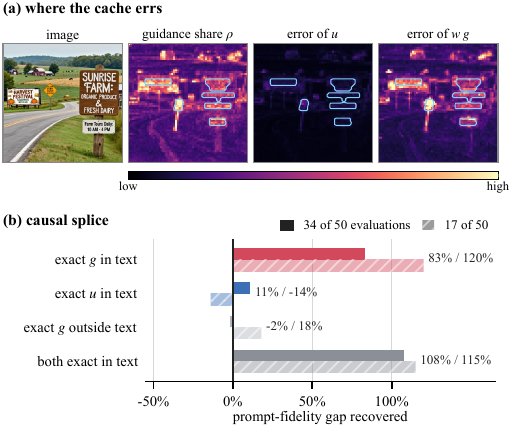}
\caption{\textbf{The text error of the cache comes mainly from the guidance term.} (a)~Guidance share $\rho$ and extrapolation errors of $u$ and $w\,g$ on one prompt. (b)~Lost prompt fidelity recovered by making one part exact in one region.}
\label{fig:split}
\end{figure}

A plausible explanation is that character strokes are high-frequency details that approximations readily lose.
To examine this explanation, we compare text with non-text regions of matched texture.
Text accumulates no more pixel error than these equally detailed regions, so pixel error alone does not explain the failure.
We instead decompose the caching error along CFG into the unconditional shared part and the guidance term, the difference between conditional and unconditional predictions (\cref{fig:split}).
The shared part is extrapolated similarly well in text and control regions.
The guidance term occupies a larger share of the output on text, its error is amplified by the guidance weight, and it is the component that caches of every order extrapolate worst on text.

To test whether this component accounts for the lost text, we intervene in the cached trajectory by making the guidance term exact within text regions.
This correction alone recovers 83\% of the lost prompt fidelity, while the same intervention outside text recovers little.
Two observations further connect the failure to guidance.
Increasing Z-Image's guidance scale from 3 to 7 more than doubles the text loss caused by caching.
Qwen-Image rescales each token's guided output, which limits guidance error, and loses about half as much text as Z-Image.
Disabling this rescaling doubles its loss.
Together, these results locate the main source of text degradation in the guidance term for these true-CFG models and establish guidance correction as a way to restore text.

Turning this finding into an acceleration method requires identifying where to spend the remaining computation.
In our comparisons, FORA and TaylorSeer lose text, while ToCa and TeaCache preserve more text at the cost of greater deviation from the full model's output.
Token-wise caches use attention scores that barely distinguish text from regions of matched texture.
CFG accelerators~\cite{castillo2023ag,kynkaanniemi2024interval,lv2025fastercache} operate on entire steps or frequency bands across the image.
These choices do not directly target the guidance error concentrated on a small set of text tokens.

We propose \method to target this error using the guidance signal already available in the cache.
At each anchor step, it stores the shared part and the guidance term separately.
On a cached step, it predicts each token's guidance error from the stored guidance terms, recomputes both CFG branches for the tokens with the largest predicted error, and extrapolates the remaining tokens as before.
The number of recomputed tokens follows the predicted error, allocating more computation to steps immediately after warm-up and farther from an anchor.
Token selection uses only the stored guidance terms and takes about half a millisecond per step.
Because the underlying extrapolation remains unchanged, \method applies to caches of any order.

We evaluate \method on LongText-Bench with Z-Image and Qwen-Image, comparing it with step reduction and reuse, forecasting, token-wise, adaptive, and CFG caching methods.
At 2.4$\times$ fewer FLOPs, it raises TaylorSeer's word accuracy on Z-Image from 46\% to 61\% and restores full-compute prompt fidelity on both models.
It matches or exceeds the compared caches in text rendering while remaining much closer to the full model's output.
The gains hold across all tested cache orders.
Combined with a 25-step schedule, \method preserves more text than 15-step sampling on both models at a higher speedup.

Our main contributions are as follows.
\begin{enumerate}
  \item We identify silent text rewriting, a failure of feature caching in which the rendered text is replaced while image-level metrics barely change, and measure it with an OCR-based protocol on two text-rendering DiTs.

  \item We trace the failure to the guidance term of CFG, show that the loss follows the guidance weight, and verify with causal interventions that correcting the guidance inside the text recovers most of the lost text.

  \item We propose \method, which recomputes only the tokens selected by the stored guidance signal. It applies to caches of any order, restores full-compute prompt fidelity on Z-Image and Qwen-Image, and with fewer sampling steps keeps more text than step reduction at fewer FLOPs.
\end{enumerate}
\section{Related Work}
\label{sec:related}

\paragraph{Efficient diffusion transformers.}
Diffusion~\cite{ho2020ddpm,song2021ddim,song2021sde,rombach2022ldm} and flow-matching models~\cite{lipman2023flow,liu2023rectified} have moved from U-Nets to large transformers~\cite{peebles2023dit,chen2024pixart,li2024hunyuandit,esser2024sd3}, and current text-to-image DiTs~\cite{wu2025qwenimage,zimage2025} run a network with billions of parameters dozens of times per image, twice per step under CFG.
Acceleration follows two routes.
The first reduces the number of steps, either with higher-order ODE solvers~\cite{lu2022dpmsolver,lu2022dpmsolverpp,zhao2023unipc} or by distilling the sampler into a few-step student~\cite{salimans2022progressive,song2023consistency,luo2023lcm,sauer2024add,yin2024dmd}, which must be trained for each model.
The second lowers the cost of each step, through quantization~\cite{li2023qdiffusion,li2025svdquant}, token merging~\cite{bolya2023tome,bolya2023tomesd}, or the reuse of computation across steps.
The two routes combine, and we show in \cref{sec:exp:main} that \method on top of fewer steps keeps more text than fewer steps alone.

\paragraph{Feature caching.}
Feature caching exploits the similarity of features at adjacent steps.
DeepCache~\cite{ma2024deepcache} and Faster Diffusion~\cite{li2024fasterdiffusion} reuse U-Net features, and FORA~\cite{selvaraju2024fora}, $\Delta$-DiT~\cite{chen2024deltadit}, and PAB~\cite{zhao2025pab} carry the idea to transformers by reusing attention, MLP, or block outputs at fixed intervals, while Learning-to-Cache~\cite{ma2024l2c} learns which layers to reuse.
TeaCache~\cite{liu2025teacache} and AdaCache~\cite{kahatapitiya2024adacache} make the schedule adaptive and refresh the cache when the output is estimated to have changed.
Token-wise methods refresh only part of the image at each step: ToCa~\cite{zou2024toca} scores tokens by attention and cache frequency, DuCa~\cite{zou2024duca} alternates aggressive and conservative caching steps, and RAS~\cite{liu2025ras} updates the regions the model currently focuses on.
Instead of reusing a stale feature, TaylorSeer~\cite{liu2025taylorseer} forecasts it with a Taylor expansion over past steps, and later work replaces the expansion with Hermite polynomials~\cite{feng2025hicache}, solvers chosen per feature group~\cite{hyca2025}, or learned stage-aware predictors~\cite{cai2026lesa} to reach higher speedups.
FoCa~\cite{zheng2025foca} pairs each forecast with a calibration step derived from an ODE view of the feature trajectory, and SpeCa~\cite{liu2025speca} verifies each forecast at low cost and computes the step fully when it is rejected.
These methods report image-level metrics~\cite{heusel2017fid,hessel2021clipscore,xu2023imagereward,zhang2018lpips}, and rendered text appears only in their qualitative figures.
None of their refresh signals reads the guidance term, where we find most of the text error (\cref{sec:mechanism}).

\paragraph{Accelerating classifier-free guidance.}
CFG~\cite{ho2022cfg} doubles the cost of each step, and several methods remove part of it.
Adaptive Guidance~\cite{castillo2023ag} and OUSAC~\cite{ousac2025} skip the unconditional pass on steps where the two branches agree, guidance intervals~\cite{kynkaanniemi2024interval} apply guidance only within a range of noise levels, and guidance distillation~\cite{meng2023guideddistill} merges both passes into one network.
CFG++~\cite{chung2025cfgpp} reformulates the guided update itself, and S-CFG~\cite{shen2024scfg} and SAMG~\cite{li2026samg} vary the guidance strength across the image at full cost.
Closest to our setting, FasterCache~\cite{lv2025fastercache} caches the difference between the two branches and reuses its low- and high-frequency bands at different rates.
These methods act on whole steps or frequency bands, whereas \method recomputes both branches on the tokens where the cached guidance is predicted to fail.

\paragraph{Visual text rendering.}
TextDiffuser~\cite{chen2023textdiffuser,chen2024textdiffuser2} plans character layouts, GlyphControl~\cite{yang2023glyphcontrol} and AnyText~\cite{tuo2024anytext} condition on rendered glyphs, and Glyph-ByT5~\cite{liu2024glyphbyt5} aligns a character-aware text encoder with glyph images.
FreeText~\cite{freetext2026} and TextGuider~\cite{textguider2025} instead steer attention toward the target characters at inference time.
Recent DiTs render long multilingual text without such modules~\cite{wu2025qwenimage,zimage2025}, and LongText-Bench~\cite{xomni2025} and CVTG-2K~\cite{cvtg2025} score it with OCR.
All assume full compute; we study text under acceleration.
\section{Text Lives in the Guidance}
\label{sec:protocol}

\subsection{Preliminaries}
\label{sec:prelim}

\paragraph{Classifier-free guidance.}
At step $t$, CFG~\cite{ho2022cfg} evaluates the unconditional prediction $u_t = f_\theta(x_t, t, \varnothing)$ and the conditional prediction $c_t = f_\theta(x_t, t, y)$, and the sampler uses
\begin{equation}
  \hat v_t = u_t + w\,g_t, \qquad g_t = c_t - u_t.
  \label{eq:cfg}
\end{equation}
We call $u_t$ the \emph{shared part}, $g_t$ the \emph{guidance term}, and $w$ the effective guidance weight ($w = 1+s$ for samplers that write $c_t + s\,g_t$, as Z-Image does); all quantities are token-wise.

\paragraph{Feature caching.}
A cache computes both branches only on \emph{anchor} steps $\mathcal{A}$ and predicts the rest, $\tilde v_t = \mathrm{Ext}_t[\hat v]$ for $t \notin \mathcal{A}$, with an order-$O$ predictor: reuse for $O{=}0$ (FORA~\cite{selvaraju2024fora}) or Taylor extrapolation for $O \geq 1$ (TaylorSeer~\cite{liu2025taylorseer}).

\subsection{Caching Silently Rewrites Text}
\label{sec:protocol:phenomenon}

\paragraph{Protocol.}
\label{sec:protocol:def}
Accelerated and full-compute samplers share noise, prompt, and guidance scale, and the full-compute image serves as the reference; a sample is \emph{on-trajectory} when its PSNR to the reference exceeds about 22\,dB.
Text is read with PP-OCR~\cite{du2020ppocr} and scored by \emph{prompt fidelity} ($1-\mathrm{NED}$ to the prompt's target strings) and \emph{word accuracy}, the share of target strings that the OCR reads back verbatim; box fidelity to the reference crop is used for analysis (supplement).

\paragraph{Silent text rewriting.}
\label{sec:protocol:compute}
On Z-Image, a TaylorSeer cache that computes 34 of 50 steps stays on the full-compute trajectory, yet it lowers word accuracy by 0.078 while ImageReward and CLIP Score barely change.
Sampling with the same 34 steps loses no text, so the loss comes from how the cache predicts the skipped steps rather than from computing less.
It grows with cache order and speedup (\cref{tab:models}) and also appears on Qwen-Image.
There, fewer sampling steps lose as much text but visibly lower ImageReward, which the cache does not (\cref{tab:main}, supplement).

\subsection{Why Text Breaks}
\label{sec:mechanism}

We analyze the on-trajectory regime of the cache above and compare text boxes with a \emph{texture-matched control}, the non-text pixels whose gradient magnitude is at least the in-box median.
Along the two CFG parts, the cache error of token $i$ on a non-anchor step is
\begin{equation}
  e_t^{(i)} = \underbrace{\mathrm{Ext}_t[u]^{(i)} - u_t^{(i)}}_{e_{u,t}^{(i)}} \;+\; w\,\underbrace{\big(\mathrm{Ext}_t[g]^{(i)} - g_t^{(i)}\big)}_{e_{g,t}^{(i)}},
  \label{eq:err}
\end{equation}
and for an order-$O$ predictor with anchors $h$ apart each part errs as $O(h^{O+1}\|\partial_\tau^{O+1}x\|)$, $x \in \{u,g\}$.
We measure where each part of this error lands; all measurements are listed in \cref{tab:supp:mech}.

\paragraph{Text is not computed more coarsely.}
In decoded pixels, text errs 3--4$\times$ more than the background but about as much as the texture-matched control (\cref{tab:supp:mech}).
What goes wrong is the identity of the glyph: a perturbation that barely changes texture can turn a stroke into a neighboring character.
Under latent noise of equal norm, text holds its box fidelity as well as the control up to a threshold and then breaks, while the control degrades gradually.

\paragraph{Text takes a larger guidance share.}
In text, the guidance term makes up a larger part of the output, with $\|w\,g\|/\|\hat v\|$ at 1.66$\times$ that of the control for English and Chinese alike.
Its error is also scaled by $w$, so the text loss of the cache grows with the guidance scale (\cref{sec:exp:analysis}).
Qwen-Image rescales each token's guided output to its conditional norm, which caps $\|w\,e_g\|$, and it loses less text under the same cache.
With the rescaling switched off, TaylorSeer on Qwen-Image loses twice as much word accuracy (0.169 vs.\ 0.084), about as much as on Z-Image.

\paragraph{Caches extrapolate the guidance term worst.}
When the two parts are extrapolated separately, the guidance term errs about twice as much in text as in the control for reuse and first order (1.3--1.4$\times$ for second order), while the shared part errs about equally in both (\cref{fig:split}a); Qwen-Image shows the same pattern (\cref{tab:supp:mech}).
In text, $g$ also becomes rougher toward the end of sampling.

\paragraph{Exact guidance restores the text.}
\label{sec:method:causal}
We test this causally by splicing exact values into the cached trajectory, one part in one region.
The values come from a full forward pass on the intervened run's own current latent; the full-compute image only provides the box mask.
Restoring the guidance inside the OCR boxes recovers 83\% of the lost prompt fidelity and 90\% of the lost word accuracy.
Restoring the shared part, or the guidance outside the text, recovers little (\cref{tab:causal}), and the same holds at 17 of 50 evaluations, where \method operates (\cref{fig:split}b).

\paragraph{The guidance share finds the text.}
The attention score that ToCa~\cite{zou2024toca} uses barely separates text from the control (AUROC 0.54--0.64).
The guidance share does: recomputing its top tokens restores box fidelity as well as recomputing the OCR boxes themselves (\cref{tab:supp:mech}), and \cref{sec:exp:ablation} compares the selectors at equal cost.

\label{sec:mechanism:remedy}
The guidance term thus carries most of the text error of the cache, and its own magnitude shows where that error lies.
\Cref{sec:method} turns this into a method that keeps caching the shared part and recomputes both branches on the tokens where the guidance error is large.
\section{\method}
\label{sec:method}

\begin{figure*}[t]
\centering
\includegraphics[width=\linewidth]{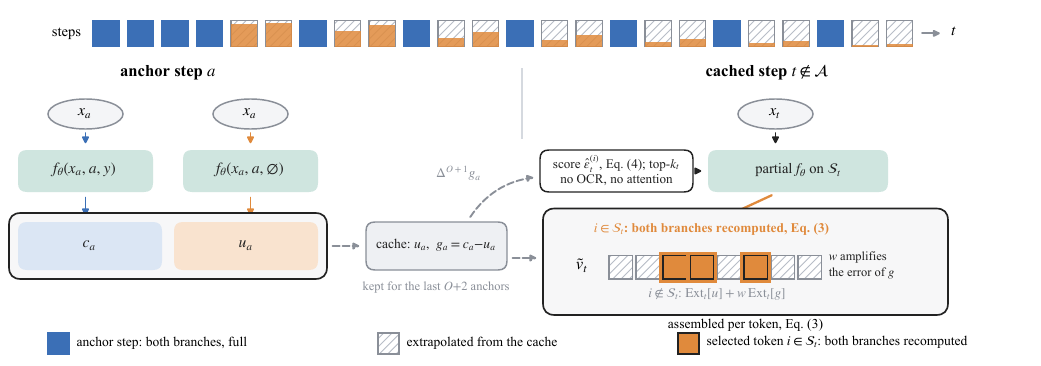}
\caption{\textbf{Overview of \method.} Anchor steps (left) store the shared part $u_a$ and the guidance term $g_a$. Cached steps (right) score tokens from the stored $g_a$, recompute both branches on the top-$k_t$ tokens, and extrapolate the rest.}
\label{fig:method}
\end{figure*}

\subsection{Split Cache with Recomputed Guidance}
\label{sec:method:overall}

At each anchor step we store $u_a$ and $g_a$ instead of $\hat v_a$.
On a non-anchor step $t$, the shared part is extrapolated for all tokens, $\tilde u_t = \mathrm{Ext}_t[u]$.
On a set $\mathcal{S}_t$ of $k_t$ tokens we recompute both branches with a partial forward pass, giving $\check u_t$ and $\check c_t$; elsewhere both parts are extrapolated:
\begin{equation}
  \tilde v_t^{(i)} =
  \begin{cases}
    \check u_t^{(i)} + w\,\big(\check c_t^{(i)} - \check u_t^{(i)}\big), & i \in \mathcal{S}_t,\\[2pt]
    \tilde u_t^{(i)} + w\,\mathrm{Ext}_t[g]^{(i)}, & i \notin \mathcal{S}_t.
  \end{cases}
  \label{eq:guidecache}
\end{equation}

\paragraph{Recomputing both branches.}
Recomputing only the conditional branch and pairing it with the extrapolated $\tilde u_t$ gives $(1-w)\,\tilde u_t^{(i)} + w\,\check c_t^{(i)}$: the guidance term is no longer extrapolated, but the shared-part error enters as $(1-w)\,e_{u,t}^{(i)}$, amplified by $w-1 = 5$ on Z-Image (\cref{sec:exp:ablation}).

Each branch on $\mathcal{S}_t$ is a ToCa-style~\cite{zou2024toca} partial forward pass in which the selected tokens are the queries and the keys and values of all other tokens, caption tokens included, are read from the last anchor.
The recomputed $\check u_t$ and $\check c_t$ thus approximate the full-compute branches $u_t$ and $c_t$, and they coincide with them at $r = 100\%$ when the caption keys and values are also recomputed every step (supplement).
\label{sec:method:analysis}With $k_t = 0$, \cref{eq:guidecache} is the standard cache; since $\mathrm{Ext}$ is untouched, \method works with caches of any order.

\subsection{Choosing the Tokens}
\label{sec:method:select}

\begin{figure}[t]
\centering
\includegraphics[width=.85\linewidth]{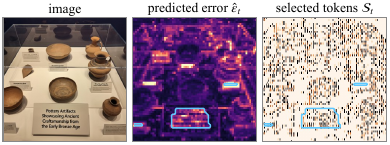}
\caption{\textbf{Tokens selected by \method} on one Z-Image prompt (34 of 50 evaluations), with the predicted guidance error $\hat\epsilon_t$ and the selection frequency. Without seeing the OCR boxes (blue), it selects 77\% of the tokens inside them and 28\% elsewhere.}
\label{fig:sel}
\end{figure}

\paragraph{Predicted guidance error.}
By \cref{eq:err}, the guidance error of token $i$ scales with $w\,h^{O+1}\|\partial^{O+1}_\tau g^{(i)}\|$.
Both factors come at no network cost: $h = t - a$ is the distance to the last anchor, and the derivative is the finite difference of the guidance terms stored at the last $O+2$ anchors, which a Taylor cache keeps anyway:
\begin{equation}
  \hat\epsilon_t^{(i)} = w\,(t - a)^{O+1}\,\big\|\Delta^{O+1} g_a^{(i)}\big\|.
  \label{eq:predict}
\end{equation}
The finite difference is noisy when anchors are far apart, so we shrink it toward its zeroth-order case $w\,\|g_a^{(i)}\|$ and rank tokens by the geometric mean $s_t^{(i)} = \big(\hat\epsilon_t^{(i)}\, w\,\|g_a^{(i)}\|\big)^{1/2}$, with equal weights fixed a priori.
On the trajectory, this shrunk score captures more of the true guidance error than either factor alone at both 13 and 17 anchors (supplement).
We select $\mathcal{S}_t = \operatorname{top}_{k_t}(s_t)$ without attention scores, OCR, or prompt text, yet the selected tokens fall on the text (\cref{fig:sel}).

\paragraph{Guidance share as the zeroth-order case.}
Before $O+2$ anchors exist, \cref{eq:predict} reduces to $w\,\|g_a^{(i)}\|$; normalized, it is the guidance share $\rho_a^{(i)} = \|w\,g_a^{(i)}\| / \|\hat v_a^{(i)}\|$\phantomsection\label{eq:selector} of \cref{sec:mechanism}, which we use during warm-up.
Both scores read only the guidance terms the cache already stores.
Most of the gain comes from this signal (\cref{sec:exp:ablation}), and $s_t$ refines it as our default.

\subsection{Error-Proportional Budget}
\label{sec:method:budget}

Given a target average ratio $r$ of recomputed tokens, the per-step budget follows the predicted error:
\begin{equation}
  k_t = \Big\lfloor r N \cdot \bar\epsilon_t \,/\, \bar\epsilon_{\text{ref}} \Big\rceil_{[0,\; r_{\max} N]},
  \label{eq:budget}
\end{equation}
where $\bar\epsilon_t$ is the mean of the top-$rN$ shrunk scores $s_t$ at step $t$ and $\bar\epsilon_{\text{ref}}$ its average over a few CVTG-2K prompts disjoint from the evaluation set, computed once per model.
A partial step costs about $r$ of a full step, and the average budget stays close to $r$ (33.5\% of the tokens per partial step at $r{=}30\%$ on Z-Image).
The budget follows the absolute predicted error, which is largest in the first anchor interval after warm-up and grows with $h$ within each interval.
\section{Experiments}
\label{sec:exp}

\begin{table*}[t]
\centering
\footnotesize
\setlength{\tabcolsep}{3.4pt}\renewcommand{\arraystretch}{0.9}
\caption{\textbf{Quantitative comparison on LongText-Bench} at $1024^2$. Methods marked $^*$ are our ports to Z-Image, and bold marks the best text score in each block.}
\label{tab:main}
\resizebox{\linewidth}{!}{\begin{tabular}{l cccccc cccccc}
\toprule
 & \multicolumn{6}{c}{Qwen-Image} & \multicolumn{6}{c}{Z-Image} \\
\cmidrule(lr){2-7}\cmidrule(lr){8-13}
Method & Word$\uparrow$ & Prompt$\uparrow$ & PSNR$\uparrow$ & LPIPS$\downarrow$ & FLOPs$\uparrow$ & Lat.$\uparrow$ & Word$\uparrow$ & Prompt$\uparrow$ & PSNR$\uparrow$ & LPIPS$\downarrow$ & FLOPs$\uparrow$ & Lat.$\uparrow$ \\
\midrule
Full (50 steps) & 0.605 & 0.615 & $\infty$ & 0 & 1.00$\times$ & 1.00$\times$ & 0.621 & 0.665 & $\infty$ & 0 & 1.00$\times$ & 1.00$\times$ \\
\midrule
\multicolumn{13}{l}{\textit{17 anchor evaluations (2.94$\times$ FLOPs)}} \\
Step reduction, 17 steps & 0.519 & 0.579 & 13.5 & 0.417 & 2.94$\times$ & 2.87$\times$ & 0.587 & 0.666 & 12.9 & 0.421 & 2.94$\times$ & 2.91$\times$ \\
Step reduction, 22 steps & 0.547 & 0.604 & 15.2 & 0.321 & 2.27$\times$ & 2.24$\times$ & 0.607 & 0.666 & 14.0 & 0.353 & 2.27$\times$ & 2.26$\times$ \\
FasterCache$^*$ & -- & -- & -- & -- & -- & -- & 0.565 & 0.652 & 15.4 & 0.327 & 1.64$\times$ & 1.70$\times$ \\
ToCa$^*$ & -- & -- & -- & -- & -- & -- & 0.604 & 0.662 & 13.3 & 0.393 & 2.32$\times$ & 2.08$\times$ \\
TeaCache$^*$ & -- & -- & -- & -- & -- & -- & 0.585 & 0.660 & 14.0 & 0.340 & 2.57$\times$ & 2.57$\times$ \\
FORA & 0.538 & 0.605 & 23.4 & 0.099 & 2.94$\times$ & 2.76$\times$ & 0.528 & 0.651 & 16.7 & 0.243 & 2.94$\times$ & 2.90$\times$ \\
\rowcolor{blue!6}
FORA + \method ($r{=}10\%$) & 0.583 & 0.616 & 24.0 & 0.093 & 2.37$\times$ & 2.08$\times$ & 0.601 & 0.666 & 16.8 & 0.237 & 2.33$\times$ & 2.20$\times$ \\
TaylorSeer & 0.521 & 0.597 & 21.2 & 0.130 & 2.94$\times$ & 2.76$\times$ & 0.464 & 0.625 & 15.0 & 0.306 & 2.94$\times$ & 2.90$\times$ \\
\rowcolor{blue!6}
TaylorSeer + \method ($r{=}10\%$) & 0.583 & \textbf{0.618} & 23.1 & 0.104 & 2.46$\times$ & 2.11$\times$ & 0.606 & \textbf{0.670} & 16.2 & 0.256 & 2.41$\times$ & 2.28$\times$ \\
\rowcolor{blue!6}
TaylorSeer + \method ($r{=}20\%$) & \textbf{0.588} & 0.617 & 24.3 & 0.083 & 2.14$\times$ & 1.91$\times$ & 0.608 & \textbf{0.670} & 16.9 & 0.229 & 2.07$\times$ & 1.97$\times$ \\
\rowcolor{blue!6}
TaylorSeer + \method ($r{=}30\%$) & 0.585 & 0.616 & 25.3 & 0.070 & 1.89$\times$ & 1.71$\times$ & \textbf{0.612} & 0.668 & 17.8 & 0.198 & 1.81$\times$ & 1.70$\times$ \\
\midrule
\multicolumn{13}{l}{\textit{13 anchor evaluations (3.85$\times$ FLOPs)}} \\
Step reduction, 13 steps & 0.460 & 0.555 & 12.4 & 0.506 & 3.85$\times$ & 3.72$\times$ & 0.546 & 0.650 & 12.1 & 0.486 & 3.85$\times$ & 3.77$\times$ \\
ToCa$^*$ & -- & -- & -- & -- & -- & -- & 0.567 & 0.658 & 12.4 & 0.458 & 2.76$\times$ & 2.47$\times$ \\
TeaCache$^*$ & -- & -- & -- & -- & -- & -- & 0.539 & 0.651 & 12.8 & 0.425 & 3.23$\times$ & 3.16$\times$ \\
FORA & 0.310 & 0.526 & 20.6 & 0.189 & 3.85$\times$ & 3.52$\times$ & 0.242 & 0.496 & 15.1 & 0.375 & 3.85$\times$ & 3.76$\times$ \\
\rowcolor{blue!6}
FORA + \method ($r{=}10\%$) & 0.488 & 0.603 & 21.5 & 0.175 & 2.94$\times$ & 2.38$\times$ & 0.499 & 0.657 & 15.0 & 0.353 & 2.84$\times$ & 2.58$\times$ \\
\rowcolor{blue!6}
FORA + \method ($r{=}20\%$) & 0.515 & 0.606 & 22.0 & 0.161 & 2.39$\times$ & 2.05$\times$ & 0.565 & \textbf{0.667} & 15.2 & 0.337 & 2.29$\times$ & 2.14$\times$ \\
\rowcolor{blue!6}
FORA + \method ($r{=}30\%$) & \textbf{0.522} & \textbf{0.613} & 22.4 & 0.148 & 2.02$\times$ & 1.76$\times$ & \textbf{0.579} & \textbf{0.667} & 15.5 & 0.314 & 1.93$\times$ & 1.80$\times$ \\
TaylorSeer & 0.124 & 0.357 & 13.7 & 0.397 & 3.85$\times$ & 3.52$\times$ & 0.057 & 0.220 & 10.3 & 0.675 & 3.85$\times$ & 3.77$\times$ \\
\rowcolor{blue!6}
TaylorSeer + \method ($r{=}10\%$) & 0.388 & 0.553 & 16.8 & 0.293 & 3.03$\times$ & 2.44$\times$ & 0.334 & 0.580 & 11.9 & 0.552 & 2.97$\times$ & 2.75$\times$ \\
\midrule
\multicolumn{13}{l}{\textit{25-step schedule (3.1--3.7$\times$ FLOPs)}} \\
Step reduction, 15 steps & 0.485 & 0.569 & 13.0 & 0.456 & 3.33$\times$ & 3.20$\times$ & 0.565 & 0.663 & 12.5 & 0.452 & 3.33$\times$ & 3.30$\times$ \\
TaylorSeer, 16 anchors & 0.526 & 0.604 & 16.2 & 0.280 & 3.12$\times$ & 2.98$\times$ & 0.551 & 0.657 & 14.4 & 0.334 & 3.12$\times$ & 3.10$\times$ \\
\rowcolor{blue!6}
TaylorSeer, 12 anchors + \method ($r{=}24\%$) & 0.552 & \textbf{0.612} & 16.2 & 0.282 & 3.26$\times$ & 3.01$\times$ & \textbf{0.605} & \textbf{0.666} & 14.2 & 0.340 & 3.14$\times$ & 2.96$\times$ \\
\rowcolor{blue!6}
TaylorSeer, 12 anchors + \method ($r{=}10\%$) & \textbf{0.554} & 0.611 & 16.1 & 0.288 & 3.71$\times$ & 3.36$\times$ & 0.599 & 0.661 & 13.9 & 0.358 & 3.63$\times$ & 3.50$\times$ \\
\bottomrule
\end{tabular}}
\end{table*}

\begin{figure*}[t]
\centering
\includegraphics[width=\linewidth]{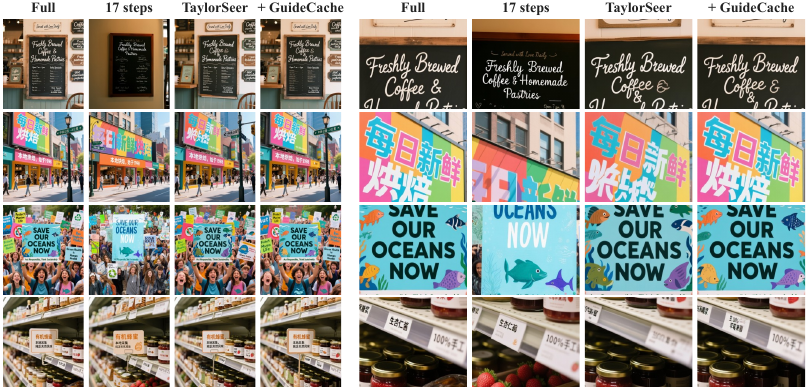}
\caption{\textbf{Qualitative comparison on Qwen-Image} for four randomly drawn prompts, with full images (left) and text crops (right). Sampling with 17 steps and TaylorSeer run at 2.94$\times$ fewer FLOPs, and \method at 2.46$\times$.}
\label{fig:qual}
\end{figure*}

\subsection{Experimental Setup}
\label{sec:exp:setup}

\paragraph{Models and benchmarks.}
We evaluate two DiTs that use true CFG and render long text at full compute, each with its default 50-step sampler and guidance scale: Z-Image~\cite{zimage2025} (6B) and Qwen-Image~\cite{wu2025qwenimage} (20B), both at $1024^2$.
Text is evaluated on LongText-Bench~\cite{xomni2025}, with all 320 English and Chinese prompts for Z-Image and a fixed 160-prompt subset for Qwen-Image, two seeds each; standard image quality is evaluated on the 200 prompts of DrawBench~\cite{saharia2022imagen}.
\Cref{tab:models,tab:ablation} and the guidance-scale and resolution studies use a fixed 48-prompt development subset of Z-Image with its own full-compute reference.
We made all design choices on this subset, and the gains on the remaining prompts match those on the full benchmark (supplement).

\paragraph{Metrics.}
Text is scored by word accuracy and prompt fidelity (\cref{sec:protocol:def}) and overall quality by ImageReward~\cite{xu2023imagereward}; following TaylorSeer and LESA, we also report PSNR and LPIPS~\cite{zhang2018lpips} to the full-compute image.
Speedup is given in FLOPs, counted from per-step token counts, and as measured latency at batch size 1.
A second OCR reader, Qwen3-VL~\cite{bai2025qwen3vl}, ranks methods alike (Kendall's $\tau = 0.72$); CLIP Score, SSIM, and other details are in the supplement.

\paragraph{Baselines.}
We compare with step reduction and reuse (FORA~\cite{selvaraju2024fora}, $O{=}0$) and forecasting (TaylorSeer~\cite{liu2025taylorseer}, $O{=}1$) caches.
We also port token-wise recomputation (ToCa~\cite{zou2024toca}), adaptive refresh (TeaCache~\cite{liu2025teacache}), and CFG caching (FasterCache~\cite{lv2025fastercache}) to Z-Image, and report CFG truncation~\cite{kynkaanniemi2024interval} in the supplement.
Two controls receive the extra compute of \method (supplement): \emph{TaylorSeer (denser)} spends it on more anchor steps, and \emph{random tokens} recomputes random tokens at the same ratio.
\method runs on FORA or TaylorSeer with 10 warm-up steps.

\subsection{Main Results}
\label{sec:exp:main}

\begin{figure}[!tb]
\centering
\includegraphics[width=\linewidth]{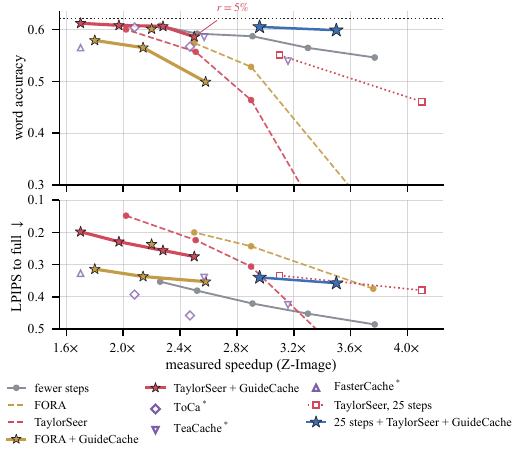}
\caption{\textbf{Speed--quality trade-off on Z-Image.} Word accuracy (top) and LPIPS (bottom) against measured speedup, with \method at $r = 5$--$30\%$ and, as blue stars, on a 25-step schedule.}
\label{fig:pareto}
\end{figure}

\paragraph{Comparison with feature caches.}
On TaylorSeer at $r{=}10\%$ and 2.4$\times$ fewer FLOPs, \method recovers 91\% of the word accuracy that the cache loses on Z-Image and 73\% on Qwen-Image, and brings prompt fidelity to the full-compute level on both models (\cref{tab:main}), with ImageReward at or above that of TaylorSeer.
On Z-Image, the extra compute alone does not explain the gain, since spending as much or more compute on random tokens, more anchors, or late anchors keeps less text (\cref{tab:supp:stack}).
Against ToCa, TeaCache, and FasterCache, \method gives higher prompt fidelity at comparable or higher speed and stays far closer to the full image in LPIPS.
Counted string by string, \method rewrites only about 8\% of the strings that full compute renders correctly on both models, fewer than any baseline (\cref{tab:supp:preserve}).
On the reuse cache, it lifts FORA from 0.528 to 0.601 word accuracy, and at 13 anchors it turns a collapsed FORA into one with near-full prompt fidelity.

\begin{table}[!tb]
\centering
\footnotesize
\setlength{\tabcolsep}{3.5pt}
\caption{\textbf{\method on caches of any order} (Z-Image development subset, prompt fidelity). FLOPs refer to the plain cache.}
\label{tab:models}
\begin{tabular}{l ccc cc}
\toprule
 & & & & \multicolumn{2}{c}{+ \method} \\
\cmidrule(lr){5-6}
Cache & Anchors & FLOPs & Plain & $r{=}20\%$ & $r{=}30\%$ \\
\midrule
Full compute & 50 & 1.0$\times$ & 0.645 & & \\
\midrule
FORA ($O{=}0$) & 28 & 1.8$\times$ & 0.641 & 0.644 & 0.643 \\
 & 22 & 2.3$\times$ & 0.630 & 0.645 & 0.654 \\
 & 17 & 2.9$\times$ & 0.622 & 0.638 & 0.654 \\
 & 13 & 3.8$\times$ & 0.475 & 0.634 & 0.645 \\
\midrule
TaylorSeer ($O{=}1$) & 28 & 1.8$\times$ & 0.624 & 0.642 & 0.648 \\
 & 17 & 2.9$\times$ & 0.601 & 0.655 & 0.647 \\
 & 13 & 3.8$\times$ & 0.213 & 0.617 & 0.637 \\
\midrule
TaylorSeer ($O{=}2$) & 28 & 1.8$\times$ & 0.577 & 0.645 & 0.647 \\
 & 22 & 2.3$\times$ & 0.506 & 0.662 & 0.652 \\
\bottomrule
\end{tabular}
\end{table}

\paragraph{Comparison with step reduction.}
Like DeepCache~\cite{ma2024deepcache}, \method keeps the text of fewer-step sampling while staying far closer to the full image.
On Z-Image it matches 22-step sampling at a similar measured speed with a much lower LPIPS (\cref{fig:pareto}), and on Qwen-Image, where fewer steps cost more text (\cref{fig:qual}), it exceeds 20-step sampling at equal FLOPs (\cref{tab:supp:stack}).
The two are also complementary.
On a 25-step schedule, TaylorSeer with \method keeps more text than 15-step sampling on both models even though it runs faster, with a word accuracy of 0.554 against 0.485 on Qwen-Image (\cref{tab:main,fig:pareto}).

\paragraph{Text loss of existing accelerators.}
Every baseline cache in \cref{tab:main} loses text: at 2.94$\times$ FLOPs, TaylorSeer lowers word accuracy from 0.621 to 0.464 on Z-Image while ImageReward barely moves (supplement), and the loss grows at 3.85$\times$.
The token-wise, adaptive, and CFG caches lose less text at their own speedups but drift farther from the full image (LPIPS 0.33--0.39 vs.\ 0.24--0.31 for FORA and TaylorSeer).

\subsection{Generality and Efficiency}
\label{sec:exp:plugin}

\begin{table}[tb]
\centering
\footnotesize
\setlength{\tabcolsep}{2.8pt}
\caption{\textbf{Ablation study} on the Z-Image development subset with TaylorSeer at 17 anchors and $r{=}30\%$. Defaults are shaded.}
\label{tab:ablation}
\begin{tabular}{llccc}
\toprule
Choice & Setting & Word$\uparrow$ & Prompt$\uparrow$ & FLOPs \\
\midrule
Plain cache & & 0.386 & 0.601 & 2.94$\times$ \\
Full compute & & 0.540 & 0.645 & 1.00$\times$ \\
\midrule
\multirow{5}{*}{Selector}
 & Random & 0.420 & 0.623 & 1.79$\times$ \\
 & ToCa attention score & 0.456 & 0.622 & 1.79$\times$ \\
 & Guidance share $\rho_a$ & 0.516 & 0.653 & 1.80$\times$ \\
 & \cellcolor{blue!6}Shrunk score $s_t$ & \cellcolor{blue!6}0.532 & \cellcolor{blue!6}0.647 & \cellcolor{blue!6}1.80$\times$ \\
 & \textcolor{gray}{OCR boxes (oracle)} & \textcolor{gray}{0.511} & \textcolor{gray}{0.655} & \textcolor{gray}{1.80$\times$} \\
\midrule
\multirow{3}{*}{\shortstack[l]{Recomputed\\branch}}
 & \cellcolor{blue!6}Both & \cellcolor{blue!6}0.532 & \cellcolor{blue!6}0.647 & \cellcolor{blue!6}1.80$\times$ \\
 & Conditional only & 0.481 & 0.651 & 2.20$\times$ \\
 & Unconditional only & 0.235 & 0.478 & 2.19$\times$ \\
\midrule
\multirow{3}{*}{Ratio $r$}
 & 10\% & 0.495 & 0.642 & 2.38$\times$ \\
 & 20\% & 0.504 & 0.655 & 2.05$\times$ \\
 & 30\% & 0.532 & 0.647 & 1.80$\times$ \\
\midrule
\multirow{2}{*}{Budget}
 & Fixed $k_t = rN$ & 0.509 & 0.651 & 1.87$\times$ \\
 & \cellcolor{blue!6}Error-proportional & \cellcolor{blue!6}0.532 & \cellcolor{blue!6}0.647 & \cellcolor{blue!6}1.80$\times$ \\
\bottomrule
\end{tabular}
\end{table}
\paragraph{Caches of any order.}
\method raises prompt fidelity over the plain cache for every order and anchor count in \cref{tab:models}, and at 30\% it stays close to full compute in every setting.
The gain is largest where the plain cache extrapolates the guidance term worst, at high order and few anchors, as for first order at 13 anchors (0.213 to 0.637).
The gain carries over to the Hermite forecaster of HiCache~\cite{feng2025hicache} (supplement).

\paragraph{Overhead.}
Token selection and the cached guidance terms cost almost nothing (0.5\,ms and 7\,MiB per step, \cref{tab:overhead}), so the cost is the partial forward pass, already included in the measured speedups of \cref{tab:main}.

\subsection{Ablation Study}
\label{sec:exp:ablation}
\label{sec:exp:exact}

\paragraph{Token selection.}
Selecting by the guidance signal that the cache stores closes most of the word-accuracy gap between the plain cache and full compute.
The guidance share reaches 0.516 and our default $s_t$, which also accounts for how fast $g$ changes, 0.532, on par with OCR boxes used as an oracle.
Random tokens and the ToCa attention score recover far less, consistent with the near-chance separation of text by attention (\cref{sec:mechanism}).
Separating the guidance term also matters, since ranking by the change of the full velocity $\hat v$ at the same budget keeps less text (supplement).

\paragraph{Recomputed branches.}
Recomputing both CFG branches gives the best text (0.532 word accuracy); the conditional branch alone reaches 0.481 and the unconditional branch alone 0.235, below the plain cache.
On the trajectory, both branches on 10\% of the tokens beat the conditional branch on 20\% in box fidelity, and the conditional branch alone drops PSNR below the plain cache through the $(1-w)\,e_u$ term (\cref{sec:method:overall}).

\paragraph{Ratio and budget.}
Even 5\% of the tokens suffice for most of the gain: at 2.64$\times$ fewer FLOPs, \method raises TaylorSeer's word accuracy on Z-Image from 0.464 to 0.586 and restores full-compute prompt fidelity on both models (\cref{tab:supp:ratio}).
Higher ratios trade speed for more text (\cref{tab:ablation}).
The error-proportional budget gives higher word accuracy than a fixed budget at nearly the same cost by moving recomputed tokens to steps where $s_t$ is large (\cref{sec:method:budget}).

\subsection{Analysis}
\label{sec:exp:analysis}
\label{sec:exp:qual}

\paragraph{Guidance scale.}
On Z-Image, \method reaches full compute at guidance scales 3 and 7, while the word-accuracy loss of first-order caching grows from 0.075 to 0.185, as the factor $w$ in \cref{eq:err} predicts.
Removing the per-token rescaling of Qwen-Image doubles the loss of the cache (\cref{sec:mechanism}) and raises the word-accuracy gain of \method at $r{=}10\%$ from 0.061 to 0.152.

\paragraph{Resolution.}
At $512^2$ each glyph covers four times fewer tokens and TaylorSeer loses far more text, yet \method recovers most of it (prompt fidelity 0.594 vs.\ 0.615 for full compute; \cref{sec:supp:gsres}).

\paragraph{Text size and script.}
\method gains most on small text, which caching rewrites most.
On Z-Image, TaylorSeer keeps a box fidelity of 0.46 on the smallest third of text and 0.81 on the largest, and \method at 2.4$\times$ adds 0.11 and 0.06 (\cref{tab:supp:strat}).
It also recovers more on Chinese than on Latin text, and Qwen-Image shows the same order.
\section{Conclusion}
\label{sec:conclusion}

A feature cache can look lossless by image-level scores and still change the words that a text-rendering DiT writes.
In the true-CFG models we studied, most of this error enters through the guidance term, whose stored values also show where the error is large.
\method recomputes both branches only on those tokens, restoring full-compute prompt fidelity at 2.4$\times$ fewer FLOPs, and with fewer sampling steps it keeps more text than step reduction at 3.6$\times$.
Rendered text is a strict and cheap test for accelerated models.
{
    \small
    \bibliographystyle{ieeenat_fullname}
    \bibliography{main}
}

\clearpage
\appendix
\twocolumn[{\centering\Large\bfseries Supplementary Material\par\vspace{1em}}]
\counterwithin*{table}{section}\counterwithin*{figure}{section}
\renewcommand{\thetable}{\thesection\arabic{table}}\renewcommand{\thefigure}{\thesection\arabic{figure}}
\renewcommand{\theHtable}{\thesection.\arabic{table}}\renewcommand{\theHfigure}{\thesection.\arabic{figure}}

\section{Implementation Details}
\label{sec:supp:impl}

\subsection{\method}

\paragraph{Partial forward pass.}
The selected image tokens act as queries, and all other tokens, including the caption tokens of the joint text--image sequence, provide keys and values from the last anchor.
If the caption tokens' keys and values are also recomputed at every step, \method at $r{=}100\%$ reproduces full compute exactly (99\,dB PSNR on Z-Image, the self-consistency of the sampler).
At $r{=}20\%$, refreshing the caption tokens costs 5\% more FLOPs and leaves prompt fidelity and LPIPS unchanged (0.637 vs.\ 0.636 and 0.220 vs.\ 0.223), so \method reads them from the anchor.

\paragraph{Selector and budget.}
Tokens are ranked by the geometric mean of the finite-difference estimate (\cref{eq:predict}) and its zeroth-order case $w\|g_a\|$, both computed per $2{\times}2$ latent patch, which is one image token.
In a same-state diagnostic on 8 Z-Image prompts at 13 and 17 anchors, this score captured the largest share of the true guidance error, and matching the anchor spacing did not help.
We use $W{=}10$ warm-up steps, $r \in \{5, 10, 20, 30\}\%$, and $r_{\max} = 3r$; $\bar\epsilon_{\text{ref}}$ is calibrated once per model and base cache on 8 CVTG-2K prompts disjoint from the benchmark.

\paragraph{Cost.}
\Cref{tab:overhead} breaks the cost of \method down per step, measured at batch size 1 for Z-Image on one RTX 4090 with CPU offload and for Qwen-Image on two RTX 4090, with memory summed over both.
Selecting the tokens and storing the guidance terms cost little next to the partial forward pass.
Full compute has no cached steps, TaylorSeer runs no network pass on them, and neither selects tokens or stores the guidance term, so these cells are left empty.

\begin{table}[!b]
\centering
\footnotesize
\setlength{\tabcolsep}{3.5pt}
\caption{\textbf{Cost per denoising step} at $1024^2$. Dashes mark parts that a method does not run.}
\label{tab:overhead}
\begin{tabular}{l ccccc}
\toprule
 & Anchor & Cached & Select & Mem.\ & Stored $g$ \\
 & (s) & (s) & (ms) & (GiB) & (GiB) \\
\midrule
\multicolumn{6}{l}{\textit{Z-Image}} \\
Full compute & 1.27 & -- & -- & 12.6 & -- \\
TaylorSeer & 1.26 & -- & -- & 12.6 & -- \\
\rowcolor{blue!6}
+ \method & 1.27 & 0.51 & 0.53 & 16.4 & 0.007 \\
\midrule
\multicolumn{6}{l}{\textit{Qwen-Image}} \\
Full compute & 1.27 & -- & -- & 39.0 & -- \\
TaylorSeer & 1.27 & -- & -- & 39.0 & -- \\
\rowcolor{blue!6}
+ \method & 1.32 & 0.49 & 0.40 & 39.2 & 0.007 \\
\bottomrule
\end{tabular}
\end{table}

\subsection{Baselines}

\paragraph{TaylorSeer and FORA.}
Our TaylorSeer cache forecasts the whole transformer output of a step from the last $O{+}1$ anchors with a Lagrange polynomial, after $W{=}10$ full warm-up steps and with one anchor every $N$ steps: $N{=}6$ gives 17 anchor evaluations (2.94$\times$) and $N{=}14$ gives 13 (3.85$\times$).
FORA is the same cache with $O{=}0$.
The denser control uses $N{=}3$, which gives 25 anchors on Z-Image (with $W{=}12$) and 24 on Qwen-Image.
The official TaylorSeer code forecasts each attention and MLP module separately.
A variant in that style, with the same anchors and the residual stream rebuilt from the current input embedding and adaLN gates, is weaker on 64 Z-Image prompts: it lowers word accuracy by 0.036 at 17 anchors ($O{=}1$), lowers prompt fidelity from 0.230 to 0.083 at 13 anchors, and raises LPIPS by 0.03--0.14 for both orders, so we report the block-level variant.

\paragraph{FasterCache, ToCa, and TeaCache.}
We run these methods on Z-Image only, since none of them has a Qwen-Image port.
FasterCache has no image-model port either, so we reimplement its CFG cache from the paper, which reuses the low- and high-frequency parts of the branch difference, with a spatial FFT split at radius 0.2, a phase boundary at half of the sampling, both weights at 0.2, CFG reuse every 5 steps after 17 warm-up steps, and attention reuse every 2 steps from step 2 with a weight rising linearly from 0 to 1; image models have no temporal attention, so the port differs from the official video code.
For ToCa, we port its token selection (attention score plus cache frequency) to the single-stream DiT with a fresh interval of $N{=}3$ in the 17-anchor block and 4 in the 13-anchor block and a cache ratio of $R{=}90\%$, so 10\% of the image tokens are recomputed in the MLP on cached steps while self-attention is reused; $\lambda_3{=}0.25$, $2{\times}2$ spatial uniformity with $\lambda_4{=}1$, and attention scores from 256 sampled queries.
For TeaCache, the indicator is the relative L1 change of the modulated input of the first transformer block, rescaled by a fourth-order polynomial fitted on 16 CVTG-2K prompts disjoint from LongText-Bench ($R^2 = 0.89$).
The threshold is set by bisection on 8 of these prompts to 17 or 13 computed steps on average ($\delta = 0.157$ and $0.231$).
Skipped steps reuse the cached residual of the 30 main blocks, the first and last steps are always computed, and since Z-Image evaluates both CFG branches in one batch, they share the skip decision.
We do not compare with LESA~\cite{cai2026lesa}, whose stage-aware predictors must be trained per model and are not released for either model.
Each of these ports needs per-model tuning (the rescaling polynomial of TeaCache, the cache schedule of ToCa, and the frequency split of FasterCache), which we carried out for Z-Image only.

\subsection{Evaluation Details}
\label{sec:supp:protocol}

\paragraph{Pairing.}
Each accelerated image is paired with the full-compute image of the same prompt, seed, and machine, and reported values average over prompts and seeds.
The machine matters because the reference itself depends on the attention kernel.
Full-compute Z-Image images from two attention kernels reach only 21.1 and 20.1\,dB PSNR against each other (48 prompts, two seeds), while caches at $\approx$2.5$\times$ differ from the reference by about 1.6\,dB.

\paragraph{Text scores.}
Text is read with PP-OCR rather than a vision-language model, which tends to repair malformed glyphs.
We use the PP-OCRv4 detector and recognizer through RapidOCR 1.4.4 with its default thresholds (binarization 0.3, box score 0.5, text score 0.5); OCR boxes are detected on the full-compute image, and boxes smaller than 64 pixels are dropped.
Word accuracy and prompt fidelity are reference-free and serve as the headline scores.
Box fidelity, the $1-\mathrm{NED}$ between the reference and accelerated crops of the same OCR box, needs a reference trajectory and is used for analysis.
The texture-matched control is the set of non-text pixels whose gradient magnitude is at least the median inside the text boxes of the same image; it covers about 1.6 times the text area.
Image quality is scored with ImageReward and CLIP Score, and closeness to full compute with PSNR and LPIPS.
As a second reader, Qwen3-VL-8B~\cite{bai2025qwen3vl} transcribes all 320 images of 16 Z-Image settings of the main comparison (seed 2000).
It scores every method higher, as expected from a reader that repairs glyphs, yet ranks the methods as PP-OCR does (Kendall's $\tau = 0.72$ over method means, per-image Pearson $r = 0.82$).
Under this reader, \method at $r{=}10\%$ still exceeds TaylorSeer by 0.032 in prompt fidelity and also leads TeaCache, FasterCache, and random tokens.

\paragraph{Development prompts.}
We made all design choices on the 48-prompt development subset of Z-Image, and the studies run only on Qwen-Image used 48 prompts of its 160-prompt subset.
On the prompts that no development run used, 272 for Z-Image and 95 for Qwen-Image, TaylorSeer with \method at $r{=}10\%$ recovers 92\% and 73\% of the word accuracy that TaylorSeer loses, against 91\% and 73\% on the full benchmark, and exceeds full compute in prompt fidelity on both models.

\paragraph{Noise floor of box fidelity.}
A generic on-trajectory error barely moves box fidelity.
Noise of 2--10\% of the latent norm added to the whole latent at step 40 lowers PSNR to 35--38\,dB but leaves box fidelity at 0.93 (24 prompts), and noise confined to the control region at 33\% of the norm (24\,dB) still gives 0.92.
The floor of the metric is therefore about 0.93, far above the 0.68 of TaylorSeer at 2.94$\times$.

\paragraph{Speed.}
FLOPs count an anchor step as a full CFG pass and a partial step as $k_t/N$ of the transformer blocks of both branches plus the unchanged refiner and final layers.
A full CFG step costs about 115 TFLOPs on Z-Image and 140 TFLOPs on Qwen-Image at $1024^2$, varying by about 2\% with the caption length.
Latency is the wall-clock time per image at batch size 1 on one A800 GPU that holds the whole model, including the anchor keys and values; it is averaged over six prompts after two warm-up prompts, and the speedup divides the full-compute time by it.
The analysis of \cref{sec:protocol} ran with CPU offload on one RTX 4090 and is therefore reported in network evaluations.

\paragraph{Models.}
We keep models whose full-compute text is good enough to measure a loss.
SD3.5 Large and CogView4 are not: their word accuracy at full compute is 0.032 and 0.061 (on 20 prompts) on LongText-Bench.

\section{Mechanism Analysis}
\label{sec:supp:mechanism}

This section collects the measurements behind \cref{sec:mechanism}.
\Cref{tab:supp:mech} lists them, and \cref{tab:causal} gives the full causal splice.

\begin{table*}[!t]
\centering
\small
\caption{\textbf{Measurements behind the analysis} in \cref{sec:mechanism}.}
\label{tab:supp:mech}
\begin{tabular}{p{0.27\linewidth} p{0.66\linewidth}}
\toprule
Quantity & Measurement \\
\midrule
Setting & Z-Image, 34 of 50 evaluations on the full-compute trajectory (23--24\,dB); anchor interval 6 for orders 0--1, 3 for order 2; 48 prompts for pixels, 12 English and 12 Chinese inside the network \\
Pixel error, text / control & 0.96--1.13$\times$ (text / background: 3--4$\times$) \\
Latent noise of equal norm & box fidelity of text within 0.02 of the control up to 17\% of the latent norm; 0.853 vs.\ 0.920 at 33\%; control SSIM 0.98 to 0.78 \\
Guidance share $\|w\,g\|/\|\hat v\|$ & text 0.50 vs.\ control 0.32 (1.66$\times$); English 1.55$\times$, Chinese 1.77$\times$ \\
Per-token rescaling (Qwen-Image) & TaylorSeer at 17 anchors loses 0.084 word accuracy and 0.018 prompt fidelity with the rescaling, and 0.169 and 0.050 without it (160 prompts, two seeds) \\
Extrapolation error of $g$, text / control & reuse 1.83$\times$, first order 1.96$\times$, second order 1.30--1.38$\times$ \\
Extrapolation error of $u$, text / control & 0.97$\times$, 1.11$\times$, 1.00--1.04$\times$ \\
Qwen-Image, first order (24 prompts) & $g$: 1.46$\times$ (interval 6), 1.36$\times$ (interval 3); $u$: 1.04$\times$, 1.08$\times$ \\
Curvature of $g$, text / control & 1.29--1.34 in steps 37--48, against 0.93--1.13 before \\
Causal splice, 34 evaluations & guidance in text 83\% / 90\% (prompt / word); shared part in text 11\% / 2\%; guidance outside text none (\cref{tab:causal}) \\
Causal splice, 17 evaluations & guidance in text 120\%; shared part $-14\%$; guidance on an equal number of control tokens 18\% \\
Attention score (ToCa-style) & AUROC text vs.\ control 0.54--0.64 \\
Guidance-share selection & box fidelity 0.811 (cache) to 0.891 (top 30\% recomputed); OCR boxes recomputed 0.890 \\
\bottomrule
\end{tabular}
\end{table*}

\begin{table}[!t]
\centering
\footnotesize
\caption{\textbf{Share of the lost text recovered by the causal splice} on Z-Image.}
\label{tab:causal}
\resizebox{\linewidth}{!}{\begin{tabular}{llccc}
\toprule
Exact part & Region & Prompt fid. & Box fid. & Word acc. \\
\midrule
Guidance term & text boxes & \textbf{83\%} & \textbf{42\%} & \textbf{90\%} \\
Shared part & text boxes & 11\% & 2\% & 2\% \\
Guidance term & control & $-2\%$ & $-2\%$ & $-6\%$ \\
\midrule
\textcolor{gray}{Both} & \textcolor{gray}{text boxes} & \textcolor{gray}{108\%} & \textcolor{gray}{45\%} & \textcolor{gray}{84\%} \\
\bottomrule
\end{tabular}}
\end{table}

\paragraph{Causal splice.}
The splice of \cref{tab:causal} runs on Z-Image at 34 of 50 evaluations over 48 prompts, where the cache trails full compute by 0.023 in prompt fidelity, 0.189 in box fidelity, and 0.096 in word accuracy.
At 17 of 50 evaluations (\cref{fig:split}b), the splice uses the 42 of these prompts whose dilated text leaves room for a control region of the same size, and a recovery above 100\% means that the spliced run scores higher than full compute.
The control splice does act on its own region: exact guidance there raises the SSIM of the control pixels to the reference from 0.812 to 0.873, while the text stays at 0.878 (0.877 for the cache, 0.946 with exact guidance in the text).
At every non-anchor step, the splice evaluates the exact transformer on the intervened run's own current latent and replaces only the masked part of the extrapolation.
It never copies values from the reference trajectory, which provides only the OCR box mask.
Box fidelity recovers less than prompt fidelity in every splice, 45\% of the gap even with both parts exact, because only the inside of the boxes is replaced while the surrounding context stays extrapolated; within that limit, exact guidance alone reaches 93\% of the recovery with both parts exact.

\begin{figure}[!t]
\centering
\includegraphics[width=\linewidth]{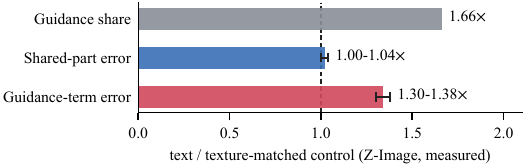}
\caption{\textbf{Cache error in text relative to the texture-matched control} on Z-Image, as the midpoint and range over cache settings.}
\label{fig:split:supp}
\end{figure}

\section{Additional Results}
\label{sec:supp:results}

\begin{table*}[!t]
\centering
\footnotesize
\setlength{\tabcolsep}{3.4pt}
\caption{\textbf{Equal-FLOPs controls, stacking, and CFG truncation}, with shaded rows repeated from \cref{tab:main}.}
\label{tab:supp:stack}
\resizebox{\linewidth}{!}{\begin{tabular}{l cccccc cccccc}
\toprule
 & \multicolumn{6}{c}{Qwen-Image} & \multicolumn{6}{c}{Z-Image} \\
\cmidrule(lr){2-7}\cmidrule(lr){8-13}
Method & Word$\uparrow$ & Prompt$\uparrow$ & PSNR$\uparrow$ & LPIPS$\downarrow$ & FLOPs$\uparrow$ & Lat.$\uparrow$ & Word$\uparrow$ & Prompt$\uparrow$ & PSNR$\uparrow$ & LPIPS$\downarrow$ & FLOPs$\uparrow$ & Lat.$\uparrow$ \\
\midrule
\multicolumn{13}{l}{\textit{Stacking on a 25-step schedule (3.1--4.2$\times$ FLOPs)}} \\
Step reduction, 15 steps & 0.485 & 0.569 & 13.0 & 0.456 & 3.33$\times$ & 3.20$\times$ & 0.565 & 0.663 & 12.5 & 0.452 & 3.33$\times$ & 3.30$\times$ \\
25 steps + TaylorSeer (12 anchors) & 0.512 & 0.597 & 15.9 & 0.295 & 4.17$\times$ & 3.90$\times$ & 0.461 & 0.620 & 13.6 & 0.380 & 4.17$\times$ & 4.10$\times$ \\
25 steps + TaylorSeer (16 anchors) & 0.526 & 0.604 & 16.2 & 0.280 & 3.12$\times$ & 2.98$\times$ & 0.551 & 0.657 & 14.4 & 0.334 & 3.12$\times$ & 3.10$\times$ \\
\rowcolor{blue!6}
25 steps + TaylorSeer (12) + \method ($r{=}10\%$) & 0.554 & 0.611 & 16.1 & 0.288 & 3.71$\times$ & 3.36$\times$ & 0.599 & 0.661 & 13.9 & 0.358 & 3.63$\times$ & 3.50$\times$ \\
\rowcolor{blue!6}
25 steps + TaylorSeer (12) + \method ($r{=}24\%$) & 0.552 & 0.612 & 16.2 & 0.282 & 3.26$\times$ & 3.01$\times$ & 0.605 & 0.666 & 14.2 & 0.340 & 3.14$\times$ & 2.96$\times$ \\
\midrule
\multicolumn{13}{l}{\textit{CFG truncation}} \\
Last 10 steps & 0.570 & 0.614 & 36.1 & 0.008 & 1.11$\times$ & 1.11$\times$ & 0.617 & 0.664 & 32.1 & 0.021 & 1.11$\times$ & 1.10$\times$ \\
Last 20 steps & 0.543 & 0.605 & 30.2 & 0.029 & 1.24$\times$ & 1.24$\times$ & 0.585 & 0.660 & 25.7 & 0.074 & 1.25$\times$ & 1.24$\times$ \\
\midrule
\multicolumn{13}{l}{\textit{Equal-FLOPs controls (1.8--3.0$\times$)}} \\
TaylorSeer (denser; 24 / 25 anchors) & 0.593 & 0.610 & 25.4 & 0.061 & 2.08$\times$ & 2.02$\times$ & 0.600 & 0.661 & 19.4 & 0.148 & 2.00$\times$ & 2.02$\times$ \\
TaylorSeer + random tokens ($30\%$) & 0.549 & 0.604 & 22.2 & 0.108 & 1.87$\times$ & 1.67$\times$ & 0.510 & 0.642 & 15.5 & 0.275 & 1.79$\times$ & 1.73$\times$ \\
TaylorSeer (13 anchors) + random tokens ($30\%$) & 0.258 & 0.475 & 16.4 & 0.289 & 2.10$\times$ & 1.80$\times$ & 0.125 & 0.384 & 11.7 & 0.562 & 1.98$\times$ & 1.86$\times$ \\
FORA, 20 anchors & -- & -- & -- & -- & -- & -- & 0.574 & 0.661 & 17.8 & 0.200 & 2.50$\times$ & 2.50$\times$ \\
TaylorSeer, 20 anchors & 0.537 & 0.609 & 23.5 & 0.086 & 2.50$\times$ & 2.40$\times$ & 0.557 & 0.658 & 16.7 & 0.224 & 2.50$\times$ & 2.51$\times$ \\
Step reduction, 20 steps & 0.538 & 0.602 & 14.6 & 0.351 & 2.50$\times$ & 2.46$\times$ & 0.592 & 0.666 & 13.6 & 0.381 & 2.50$\times$ & 2.52$\times$ \\
TaylorSeer, late-dense, 21 anchors & -- & -- & -- & -- & -- & -- & 0.561 & 0.651 & 15.0 & 0.287 & 2.38$\times$ & 2.38$\times$ \\
TaylorSeer, late-dense, 23 anchors & -- & -- & -- & -- & -- & -- & 0.576 & 0.653 & 15.1 & 0.285 & 2.17$\times$ & 2.17$\times$ \\
\rowcolor{blue!6}
TaylorSeer + \method ($r{=}10\%$) & 0.583 & 0.618 & 23.1 & 0.104 & 2.46$\times$ & 2.11$\times$ & 0.606 & 0.670 & 16.2 & 0.256 & 2.41$\times$ & 2.28$\times$ \\
\rowcolor{blue!6}
TaylorSeer + \method ($r{=}20\%$) & 0.588 & 0.617 & 24.3 & 0.083 & 2.14$\times$ & 1.91$\times$ & 0.608 & 0.670 & 16.9 & 0.229 & 2.07$\times$ & 1.97$\times$ \\
\rowcolor{blue!6}
TaylorSeer (13 anchors) + \method ($r{=}10\%$) & 0.388 & 0.553 & 16.8 & 0.293 & 3.03$\times$ & 2.44$\times$ & 0.334 & 0.580 & 11.9 & 0.552 & 2.97$\times$ & 2.75$\times$ \\
\bottomrule
\end{tabular}}
\end{table*}

\begin{table*}[!t]
\centering
\footnotesize
\setlength{\tabcolsep}{3.4pt}
\caption{\textbf{Effect of the recomputed-token ratio} on TaylorSeer at 17 anchors, with the 10--30\% rows repeated from \cref{tab:main}.}
\label{tab:supp:ratio}
\resizebox{\linewidth}{!}{\begin{tabular}{l cccccc cccccc}
\toprule
 & \multicolumn{6}{c}{Qwen-Image} & \multicolumn{6}{c}{Z-Image} \\
\cmidrule(lr){2-7}\cmidrule(lr){8-13}
Method & Word$\uparrow$ & Prompt$\uparrow$ & PSNR$\uparrow$ & LPIPS$\downarrow$ & FLOPs$\uparrow$ & Lat.$\uparrow$ & Word$\uparrow$ & Prompt$\uparrow$ & PSNR$\uparrow$ & LPIPS$\downarrow$ & FLOPs$\uparrow$ & Lat.$\uparrow$ \\
\midrule
TaylorSeer & 0.521 & 0.597 & 21.2 & 0.130 & 2.94$\times$ & 2.76$\times$ & 0.464 & 0.625 & 15.0 & 0.306 & 2.94$\times$ & 2.90$\times$ \\
\rowcolor{blue!6}
+ \method\ 5\% & 0.571 & 0.615 & 22.3 & 0.115 & 2.68$\times$ & 2.22$\times$ & 0.586 & 0.664 & 15.7 & 0.275 & 2.64$\times$ & 2.50$\times$ \\
\rowcolor{blue!6}
+ \method\ 10\% & 0.583 & 0.618 & 23.1 & 0.104 & 2.46$\times$ & 2.11$\times$ & 0.606 & 0.670 & 16.2 & 0.256 & 2.41$\times$ & 2.28$\times$ \\
\rowcolor{blue!6}
+ \method\ 20\% & 0.588 & 0.617 & 24.3 & 0.083 & 2.14$\times$ & 1.91$\times$ & 0.608 & 0.670 & 16.9 & 0.229 & 2.07$\times$ & 1.97$\times$ \\
\rowcolor{blue!6}
+ \method\ 30\% & 0.585 & 0.616 & 25.3 & 0.070 & 1.89$\times$ & 1.71$\times$ & 0.612 & 0.668 & 17.8 & 0.198 & 1.81$\times$ & 1.70$\times$ \\
\midrule
Full (50 steps) & 0.605 & 0.615 & $\infty$ & 0 & 1.00$\times$ & 1.00$\times$ & 0.621 & 0.665 & $\infty$ & 0 & 1.00$\times$ & 1.00$\times$ \\
\bottomrule
\end{tabular}}
\end{table*}

\begin{table}[!t]
\centering
\footnotesize
\setlength{\tabcolsep}{3pt}
\caption{\textbf{Rewritten and newly correct strings} in the 17-anchor block of \cref{tab:main}, in percent of the strings that full compute renders correctly or misses.}
\label{tab:supp:preserve}
\begin{tabular}{l cc cc}
\toprule
 & \multicolumn{2}{c}{Qwen-Image} & \multicolumn{2}{c}{Z-Image} \\
\cmidrule(lr){2-3}\cmidrule(lr){4-5}
Method & Rewr.$\downarrow$ & New$\uparrow$ & Rewr.$\downarrow$ & New$\uparrow$ \\
\midrule
Step reduction, 17 steps & 22.2 & 14.3 & 14.2 & 16.8 \\
Step reduction, 22 steps & 15.6 & 11.8 & 10.4 & 14.4 \\
FasterCache$^*$ & -- & -- & 13.3 & 8.2 \\
ToCa$^*$ & -- & -- & 12.1 & 17.2 \\
TeaCache$^*$ & -- & -- & 12.4 & 12.8 \\
FORA & 14.8 & 6.0 & 17.5 & 6.3 \\
\rowcolor{blue!6}
FORA + \method ($r{=}10\%$) & \textbf{7.4} & 6.6 & 8.7 & 10.8 \\
TaylorSeer & 17.5 & 6.8 & 27.7 & 5.8 \\
\rowcolor{blue!6}
TaylorSeer + \method ($r{=}10\%$) & 8.3 & 6.8 & \textbf{8.2} & 10.0 \\
\bottomrule
\end{tabular}
\end{table}

\begin{table}[!t]
\centering
\footnotesize
\setlength{\tabcolsep}{3pt}
\caption{\textbf{\method on HiCache} at 17 anchor evaluations on Z-Image.}
\label{tab:supp:hicache}
\resizebox{\linewidth}{!}{\begin{tabular}{l cccccc}
\toprule
Method & Word$\uparrow$ & Prompt$\uparrow$ & PSNR$\uparrow$ & LPIPS$\downarrow$ & FLOPs$\uparrow$ & Rewr.$\downarrow$ \\
\midrule
TaylorSeer & 0.464 & 0.625 & 15.0 & 0.306 & 2.94$\times$ & 27.7 \\
\rowcolor{blue!6}
+ \method ($r{=}10\%$) & \textbf{0.606} & 0.670 & 16.2 & 0.256 & 2.41$\times$ & \textbf{8.2} \\
HiCache & 0.491 & 0.639 & 15.2 & 0.298 & 2.94$\times$ & 22.7 \\
\rowcolor{blue!6}
+ \method ($r{=}10\%$) & 0.596 & \textbf{0.672} & 16.4 & 0.255 & 2.49$\times$ & 9.7 \\
\midrule
Full (50 steps) & 0.621 & 0.665 & $\infty$ & 0 & 1.00$\times$ & -- \\
\bottomrule
\end{tabular}}
\end{table}

\begin{table}[!t]
\centering
\footnotesize
\setlength{\tabcolsep}{3pt}
\caption{\textbf{Sensitivity to the shrinkage weight $\lambda$ and the warm-up length $W$} on the Z-Image development subset at $r{=}10\%$. Defaults are shaded.}
\label{tab:supp:sens}
\begin{tabular}{ll cccc}
\toprule
 & Setting & Word$\uparrow$ & Prompt$\uparrow$ & LPIPS$\downarrow$ & FLOPs$\uparrow$ \\
\midrule
\multirow{5}{*}{$\lambda$} & 0 ($w\|g_a\|$ only) & 0.508 & 0.629 & 0.259 & 2.30$\times$ \\
 & 0.25 & 0.508 & 0.621 & 0.252 & 2.33$\times$ \\
 & \cellcolor{blue!6}0.5 & \cellcolor{blue!6}0.521 & \cellcolor{blue!6}0.632 & \cellcolor{blue!6}0.251 & \cellcolor{blue!6}2.41$\times$ \\
 & 0.75 & 0.465 & 0.631 & 0.251 & 2.50$\times$ \\
 & 1 ($\hat\epsilon_t$ only) & 0.471 & 0.638 & 0.259 & 2.58$\times$ \\
\midrule
\multirow{6}{*}{$W$} & 6, TaylorSeer & 0.442 & 0.616 & 0.286 & 2.94$\times$ \\
 & 6, + \method & 0.509 & 0.645 & 0.263 & 2.35$\times$ \\
 & \cellcolor{blue!6}10, TaylorSeer & \cellcolor{blue!6}0.385 & \cellcolor{blue!6}0.587 & \cellcolor{blue!6}0.294 & \cellcolor{blue!6}2.94$\times$ \\
 & \cellcolor{blue!6}10, + \method & \cellcolor{blue!6}0.521 & \cellcolor{blue!6}0.632 & \cellcolor{blue!6}0.251 & \cellcolor{blue!6}2.41$\times$ \\
 & 14, TaylorSeer & 0.040 & 0.312 & 0.598 & 2.94$\times$ \\
 & 14, + \method & 0.400 & 0.604 & 0.460 & 2.40$\times$ \\
\midrule
 & Full (50 steps) & 0.557 & 0.636 & 0 & 1.00$\times$ \\
\bottomrule
\end{tabular}
\end{table}

\begin{table*}[!t]
\centering
\footnotesize
\setlength{\tabcolsep}{4.5pt}
\caption{\textbf{Per-seed prompt fidelity} for \cref{tab:models} (seeds 2000\,/\,3000\,/\,4000).}
\label{tab:supp:models}
\begin{tabular}{l ccc c cc}
\toprule
 & & & & & \multicolumn{2}{c}{+ \method} \\
\cmidrule(lr){6-7}
Cache & Anchors & FLOPs & PSNR & Plain & $r{=}20\%$ & $r{=}30\%$ \\
\midrule
Full compute & 50 & 1.0$\times$ & $\infty$ & 0.636\,/\,0.653\,/\,0.634 & & \\
\midrule
FORA ($O{=}0$) & 28 & 1.8$\times$ & 23.9--24.3 & 0.635\,/\,0.648\,/\,0.640 & 0.638\,/\,0.650\,/\,0.631 & 0.637\,/\,0.649\,/\,0.632 \\
 & 22 & 2.3$\times$ & 19.7--19.9 & 0.617\,/\,0.643\,/\,0.632 & 0.629\,/\,0.661\,/\,0.642 & 0.646\,/\,0.661\,/\,0.640 \\
 & 17 & 2.9$\times$ & 16.0--16.3 & 0.632\,/\,0.613\,/\,0.608 & 0.631\,/\,0.645\,/\,0.670 & 0.648\,/\,0.660\,/\,0.659 \\
 & 13 & 3.8$\times$ & 14.8--14.9 & 0.472\,/\,0.478\,/\,0.445 & 0.612\,/\,0.655\,/\,0.627 & 0.624\,/\,0.666\,/\,0.633 \\
\midrule
TaylorSeer ($O{=}1$) & 28 & 1.8$\times$ & 22.0--22.5 & 0.613\,/\,0.635\,/\,0.639 & 0.636\,/\,0.647\,/\,0.633 & 0.629\,/\,0.667\,/\,0.641 \\
 & 17 & 2.9$\times$ & 14.4--14.7 & 0.587\,/\,0.615\,/\,0.572 & 0.636\,/\,0.674\,/\,0.649 & 0.621\,/\,0.673\,/\,0.668 \\
 & 13 & 3.8$\times$ & 10.1--10.3 & 0.182\,/\,0.245\,/\,0.198 & 0.607\,/\,0.628\,/\,0.637 & 0.634\,/\,0.641\,/\,0.627 \\
\midrule
TaylorSeer ($O{=}2$) & 28 & 1.8$\times$ & 18.1--18.7 & 0.575\,/\,0.579\,/\,0.576 & 0.634\,/\,0.656\,/\,0.639 & 0.629\,/\,0.666\,/\,0.643 \\
 & 22 & 2.3$\times$ & 14.4--14.6 & 0.507\,/\,0.504\,/\,0.509 & 0.646\,/\,0.678\,/\,0.653 & 0.645\,/\,0.659\,/\,0.643 \\
\bottomrule
\end{tabular}
\end{table*}

\begin{table*}[!t]
\centering
\small
\caption{\textbf{Text fidelity by stratum} for TaylorSeer at 17 anchors (TS) and with \method at $r{=}10\%$.}
\label{tab:supp:strat}
\resizebox{\linewidth}{!}{\begin{tabular}{ll ccccccc ccccccc}
\toprule
 & & \multicolumn{7}{c}{Z-Image} & \multicolumn{7}{c}{Qwen-Image} \\
\cmidrule(lr){3-9}\cmidrule(lr){10-16}
 & & & \multicolumn{3}{c}{Box fidelity} & \multicolumn{3}{c}{Prompt fidelity} & & \multicolumn{3}{c}{Box fidelity} & \multicolumn{3}{c}{Prompt fidelity} \\
\cmidrule(lr){4-6}\cmidrule(lr){7-9}\cmidrule(lr){11-13}\cmidrule(lr){14-16}
Stratum & Group & $n$ & TS & +\method & $\Delta$ & TS & +\method & $\Delta$ & $n$ & TS & +\method & $\Delta$ & TS & +\method & $\Delta$ \\
\midrule
\multirow{3}{*}{Font size} & small & 213 & 0.462 & 0.574 & +0.113 & 0.401 & 0.469 & +0.068 & 106 & 0.591 & 0.646 & +0.055 & 0.486 & 0.517 & +0.031 \\
 & medium & 213 & 0.757 & 0.813 & +0.056 & 0.708 & 0.736 & +0.028 & 106 & 0.702 & 0.751 & +0.049 & 0.608 & 0.625 & +0.018 \\
 & large & 213 & 0.813 & 0.868 & +0.055 & 0.770 & 0.809 & +0.040 & 107 & 0.727 & 0.742 & +0.015 & 0.701 & 0.716 & +0.015 \\
\midrule
\multirow{2}{*}{Script} & Latin & 319 & 0.671 & 0.743 & +0.073 & 0.643 & 0.678 & +0.035 & 160 & 0.691 & 0.724 & +0.033 & 0.597 & 0.610 & +0.013 \\
 & Chinese & 320 & 0.684 & 0.760 & +0.077 & 0.610 & 0.665 & +0.055 & 159 & 0.656 & 0.702 & +0.046 & 0.600 & 0.630 & +0.030 \\
\midrule
\multirow{2}{*}{Length} & short & 319 & 0.667 & 0.727 & +0.060 & 0.623 & 0.661 & +0.037 & 159 & 0.660 & 0.689 & +0.029 & 0.557 & 0.579 & +0.023 \\
 & long & 320 & 0.688 & 0.777 & +0.089 & 0.629 & 0.682 & +0.053 & 160 & 0.687 & 0.737 & +0.050 & 0.640 & 0.660 & +0.020 \\
\midrule
\multirow{8}{*}{Carrier} & sign & 80 & 0.617 & 0.699 & +0.082 & 0.638 & 0.710 & +0.072 & 80 & 0.755 & 0.792 & +0.037 & 0.666 & 0.673 & +0.008 \\
 & label & 79 & 0.208 & 0.359 & +0.151 & 0.225 & 0.310 & +0.085 & 79 & 0.569 & 0.611 & +0.041 & 0.412 & 0.433 & +0.020 \\
 & print & 80 & 0.559 & 0.606 & +0.047 & 0.540 & 0.607 & +0.067 & 80 & 0.659 & 0.688 & +0.029 & 0.607 & 0.621 & +0.014 \\
 & web page & 80 & 0.679 & 0.751 & +0.072 & 0.735 & 0.741 & +0.007 & 80 & 0.709 & 0.760 & +0.051 & 0.706 & 0.749 & +0.042 \\
 & slide & 80 & 0.786 & 0.847 & +0.062 & 0.634 & 0.668 & +0.034 & -- & -- & -- & -- & -- & -- & -- \\
 & poster & 80 & 0.890 & 0.931 & +0.041 & 0.873 & 0.897 & +0.023 & -- & -- & -- & -- & -- & -- & -- \\
 & caption & 80 & 0.808 & 0.880 & +0.072 & 0.618 & 0.661 & +0.043 & -- & -- & -- & -- & -- & -- & -- \\
 & dialogue & 80 & 0.864 & 0.936 & +0.073 & 0.741 & 0.772 & +0.030 & -- & -- & -- & -- & -- & -- & -- \\
\bottomrule
\end{tabular}}
\end{table*}

\begin{table}[!t]
\centering
\small
\caption{\textbf{Results on DrawBench} for Z-Image, with PSNR, SSIM, and LPIPS measured against the 50-step image.}
\label{tab:supp:drawbench}
\resizebox{\linewidth}{!}{\begin{tabular}{l cc ccc c}
\toprule
Method & IR$\uparrow$ & CLIP$\uparrow$ & PSNR$\uparrow$ & SSIM$\uparrow$ & LPIPS$\downarrow$ & FLOPs \\
\midrule
Full (50 steps) & 0.936 & 32.30 & $\infty$ & 1 & 0 & 1.00$\times$ \\
TaylorSeer ($O{=}1$) & 0.763 & 31.90 & 17.6 & 0.715 & 0.315 & 2.94$\times$ \\
+ \method & 0.888 & 32.38 & 20.7 & 0.783 & 0.219 & 1.98$\times$ \\
\bottomrule
\end{tabular}}
\end{table}

\subsection{Equal-FLOPs Controls, Stacking, and CFG Truncation}

\Cref{tab:supp:stack} groups three comparisons.
The equal-FLOPs controls run on LongText-Bench as in \cref{tab:main} and spend the extra compute of \method elsewhere: on more anchors (24 on Qwen-Image, 25 on Z-Image, matching \method at $r{=}20\%$), on anchors packed into the last steps (one every 6 steps until step 38 or 32 and every 2 after, 21 or 23 anchors), or on random tokens at the same ratio.
In the 13-anchor block, the matched control is TaylorSeer at 17 anchors.
On 48 Qwen-Image prompts at 17 anchors and $r{=}20\%$, with identical per-image FLOPs, guidance-selected tokens also reach a higher box fidelity than random tokens (0.827 vs.\ 0.802).
Ranking tokens instead by the change of the full velocity $\hat v$, measured as $h^{O+1}$ times the norm of its divided difference over the last $O{+}2$ anchors, with the same budget $k_t$ and matching FLOPs, lowers the word accuracy of Z-Image at $r{=}10\%$ from 0.606 to 0.591, and the share of strings that full compute renders correctly but the cache rewrites rises from 8.2\% to 9.7\%.
The stacking runs combine a 25-step schedule with TaylorSeer at 12 anchors and \method on the full benchmark, at $r{=}10\%$ and at the ratio that matches the FLOPs of TaylorSeer at 16 anchors.
CFG truncation drops the unconditional branch in the last steps; since this saves at most half a step, it cannot reach the speeds of \cref{tab:main}.
Stacking latency is measured as in \cref{tab:main}.
The matched budget of \method is $r{=}24.17\%$ on Z-Image and $23.56\%$ on Qwen-Image, chosen to match the FLOPs of TaylorSeer at 16 anchors, with $\bar\epsilon_{\text{ref}}$ calibrated for the 25-step schedule.

\subsection{Recomputed-Token Ratio}
\label{sec:supp:ratio}

Word accuracy saturates between 10\% and 30\% while PSNR and LPIPS keep improving, and 5\% already restores prompt fidelity (\cref{tab:supp:ratio}).
On Qwen-Image, the partial pass has a larger fixed cost, so 5\% runs only slightly faster than 10\% (2.22$\times$ vs.\ 2.11$\times$ measured).

\subsection{Rewritten Strings}
\label{sec:supp:rewrite}

Word accuracy counts how many target strings are read correctly, but not which ones change.
\Cref{tab:supp:preserve} compares each method with full compute string by string, with the reader and matching of word accuracy and both seeds pooled.
Rewr.\ is the share of the strings that full compute renders correctly but the method does not, and New is the share of the strings that full compute misses but the method renders.
\method rewrites the fewest strings on both models; step reduction and ToCa reach a similar word accuracy partly by also rendering more of the strings that full compute misses.

\subsection{Newer Forecaster}
\label{sec:supp:hicache}

HiCache~\cite{feng2025hicache} replaces the Taylor basis of TaylorSeer with scaled Hermite polynomials.
We run it with its FLUX setting ($O{=}2$, $\sigma{=}0.5$) at the 17 anchors of \cref{tab:main}, plain and with \method at $r{=}10\%$, whose $\bar\epsilon_{\text{ref}}$ is calibrated for this base as in \cref{sec:method:budget}.
HiCache keeps more text than TaylorSeer but still rewrites 23\% of the strings that full compute renders correctly (\cref{tab:supp:hicache}).
\method recovers 81\% of its lost word accuracy, lifts prompt fidelity above full compute, and lowers LPIPS from 0.298 to 0.255.
SpeCa~\cite{liu2025speca} is not included, since it recomputes every rejected forecast and its cost therefore depends on the prompt.
FoCa~\cite{zheng2025foca} is not included either, as its implementation was not public when we ran these experiments.

\subsection{Shrinkage Weight and Warm-up}
\label{sec:supp:sens}

The shrunk score weights the finite-difference estimate and its zeroth-order case equally, $s_t = \hat\epsilon_t^{\lambda}\,(w\|g_a\|)^{1-\lambda}$ with $\lambda{=}0.5$ fixed a priori.
On the development subset at $r{=}10\%$, with the runs and full-compute references of \cref{tab:main} and each setting with its own $\bar\epsilon_{\text{ref}}$, $\lambda{=}0.5$ gives the highest word accuracy of five weights (\cref{tab:supp:sens}); smaller weights cost little, while larger ones spend less compute and keep less text.
With the anchor count fixed at 17, \method improves TaylorSeer at every warm-up length and recovers most of the text even at $W{=}14$, where the plain cache collapses.

\subsection{Per-Seed Results}

\Cref{tab:supp:models} lists the per-seed values behind the seed means of \cref{tab:models} (Z-Image, 48 development prompts).
Seed 4000 runs on a second machine with its own full-compute references and is not pooled.

\subsection{Stratified Results}
\label{sec:supp:strat}

\Cref{tab:supp:strat} splits text fidelity by font size, script, text length, and carrier.
Font size is split into tertiles of the mean OCR box area of the full-compute image, $n$ counts images, and $\Delta$ is computed from unrounded means.
On both models the gain of \method is largest on small text, and on Z-Image it is largest on labels, the carrier the cache renders worst.

\subsection{Standard Prompts}
\label{sec:supp:drawbench}

On DrawBench (200 prompts, one seed), \method raises the ImageReward and CLIP Score of TaylorSeer and brings the image closer to full compute (\cref{tab:supp:drawbench}).
Its budget spends fewer tokens on these prompts than on LongText-Bench (1.98$\times$ vs.\ 1.81$\times$ FLOPs at $r{=}30\%$).

\subsection{Guidance Scale and Resolution}
\label{sec:supp:limits}
\label{sec:supp:gsres}

\paragraph{Guidance scale.}
From guidance scale 3 to 7 on Z-Image (48 development prompts, 17 anchors), the word-accuracy loss of the first-order cache grows from 0.075 to 0.185 and the gain of \method from 0.118 to 0.182.
\method reaches full compute at both scales (word accuracy 0.511 vs.\ 0.469 at scale 3, 0.523 vs.\ 0.526 at 7).
Without the per-token rescaling of Qwen-Image (160 prompts, 17 anchors), full compute keeps its word accuracy (0.609 vs.\ 0.605), while TaylorSeer loses 0.169 instead of 0.084.
\method at $r{=}10\%$ recovers 0.152 instead of 0.061 and ends 0.017 below full compute (0.022 with the rescaling).

\paragraph{Resolution.}
At $512^2$ each glyph covers four times fewer tokens, and caching rewrites far more text.
On the same 48 development prompts, TaylorSeer at 17 anchors lowers prompt fidelity by 0.173 at $512^2$ against 0.049 at $1024^2$ (word accuracy $-0.220$ vs.\ $-0.172$).
\method ($r{=}20\%$, with $\bar\epsilon_{\text{ref}}$ recalibrated at $512^2$) recovers most of it, reaching 0.594 against 0.615 for full compute at $512^2$ and 0.636 against 0.636 at $1024^2$.

\section{Limitations}
\label{sec:supp:limitations}

\method relies on true CFG, which evaluates the conditional and unconditional branches separately, so it does not apply to guidance-distilled models.
Text is scored by OCR readers, which check the characters but not the typography.

\section{Additional Qualitative Results}
\label{sec:supp:qual}

\begin{figure*}[!t]
\centering
\includegraphics[width=\linewidth]{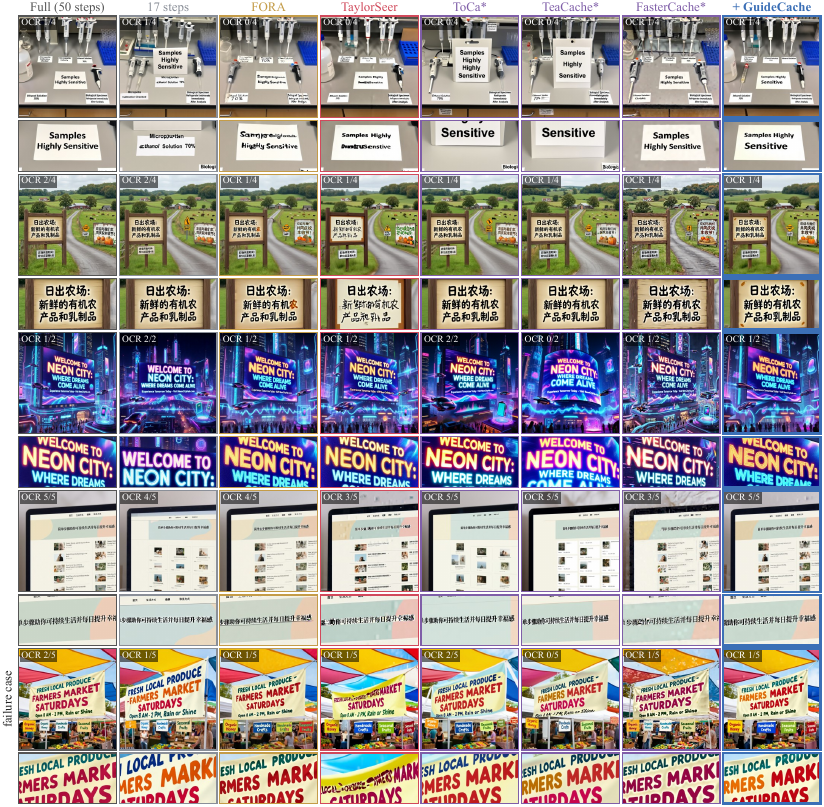}
\caption{\textbf{Randomly drawn Z-Image prompts}, full images with text crops below, and one failure case in the last row.}
\label{fig:supp:qualz}
\end{figure*}

\Cref{fig:supp:qualz} and \cref{fig:qual} of the main paper show prompts drawn at random with a fixed generator before any image was inspected.
They use the 17-anchor setting of \cref{tab:main} with \method at $r{=}10\%$, and each image is tagged with the number of target strings that PP-OCR reads verbatim.
The crops are centred on the largest text box of the full-compute image.
The last Z-Image row is a failure case: the first prompt, in a fixed random order, where \method reads fewer strings than full compute and no more than TaylorSeer.
Its strings include three small stall labels, the size at which caching rewrites most (\cref{tab:supp:strat}), and \method reads one of five, as TaylorSeer does, against two for full compute.

\end{document}